\documentclass[11pt]{article}

\usepackage[a4paper,margin=1in]{geometry}
\usepackage[T1]{fontenc}
\usepackage[utf8]{inputenc}
\usepackage{lmodern}

\usepackage{graphicx}
\usepackage{amsmath,amssymb}
\usepackage{booktabs}
\usepackage{array}
\usepackage{url}
\usepackage{enumitem}
\usepackage{tikz}
\usepackage{algorithm}
\usepackage{algpseudocode}
\usepackage[numbers]{natbib}
\title{Node Learning: A Framework for Adaptive, Decentralised and Collaborative Network Edge AI}
\author{
Eiman Kanjo\\
Nottingham Trent University, Nottingham, UK\\
Imperial College London, London, UK\\
\texttt{eiman.kanjo@ntu.ac.uk}
\and
Mustafa Aslanov\\
Nottingham Trent University, Nottingham, UK
}
\date{} 

\begin{document}

\maketitle

\begin{abstract}
The expansion of AI toward the edge increasingly exposes the cost and fragility of centralised intelligence. Data transmission, latency, energy consumption, and dependence on large data centres create bottlenecks that scale poorly across heterogeneous, mobile, and resource-constrained environments. In this paper, we introduce Node Learning, a decentralised learning paradigm in which intelligence resides at individual edge nodes and expands through selective peer interaction. Nodes learn continuously from local data, maintain their own model state, and exchange learned knowledge opportunistically when collaboration is beneficial. Learning propagates through overlap and diffusion rather than global synchronisation or central aggregation. It unifies autonomous and cooperative behaviour within a single abstraction and accommodates heterogeneity in data, hardware, objectives, and connectivity. This concept paper develops the conceptual foundations of this paradigm, contrasts it with existing decentralised approaches, and examines implications for communication, hardware, trust, and governance. Node Learning does not discard existing paradigms, but places them within a broader decentralised perspective.
\end{abstract}

\maketitle
\section{Introduction}

The increasing reliance on centralised AI has exposed structural costs that become difficult to
justify at the edge. Continuous data transmission, reliance on large data centres, and tight
coupling to wide-area connectivity introduce latency, energy overhead, and operational fragility.
As learning moves closer to where data are generated, these constraints are no longer secondary
considerations but dominant design factors.

Existing decentralised learning approaches address parts of this shift, yet retain assumptions
that limit their applicability under realistic edge conditions. Distributed learning assumes a
shared optimisation objective, coordinated execution, and relatively stable infrastructure.
Federated learning reduces raw-data centralisation, but preserves hierarchical orchestration,
round-based aggregation, and sensitivity to stragglers and non-IID data
\cite{kairouz2021federated,baccour2022pervasive}. Decentralised variants relax central coordination
through peer-to-peer exchange, but often still frame nodes as contributors to a single shared task,
rather than as persistent learners embedded in distinct contexts \cite{xu2024aiot}.

At the edge, nodes rarely operate as interchangeable workers. Devices differ in sensing modality,
compute capability, energy availability, mobility, and local objectives. Learning unfolds
continuously rather than in isolated training rounds, and collaboration is intermittent,
asymmetric, and context dependent. These characteristics suggest a shift in perspective: from
learning as a coordinated optimisation process to learning as an evolving system of interacting
entities.

This concept paper adopts that perspective through \textbf{Node Learning}. In Node Learning, each node
maintains its own learning state, adapts continuously to local observations, and cooperates with
peers opportunistically when collaboration adds value. Knowledge exchange is not restricted to
full parameter averaging and need not imply a shared global model. Nodes may exchange features,
embeddings, adapters, partial updates, or confidence signals, integrating external information in
ways shaped by context such as energy, connectivity, trust, and task relevance.

Viewed through this lens, behaviours commonly treated as distinct paradigms—federated learning,
distributed optimisation, collaborative inference—appear as operating regimes within a broader
continuum. Nodes may act independently, form transient coalitions, share resources, or propagate
learned knowledge across overlapping neighbourhoods, without relying on persistent coordination
or a single optimisation objective. Learning progresses through diffusion and accumulation rather
than convergence to a centrally defined solution.
This architectural view has implications beyond algorithms. Communication, computation, and
learning become tightly coupled; hardware heterogeneity and energy constraints shape optimisation;
and trust and governance emerge as system-level properties rather than external controls. Recent
work on gossip-based exchange, TinyML collaboration, and resource-aware edge AI illustrates the
feasibility of such interaction under severe constraints \cite{bao2025bilayergossip}, motivating
a more general abstraction.
\paragraph{Scope.}
The goal is conceptual and architectural: to establish definitions, abstractions, and design
axes that organise a fragmented literature and guide subsequent implementations. Algorithmic
instantiations and empirical validation are discussed where helpful, but formal convergence guarantees and exhaustive benchmarking are treated as future work.
Node Learning is articulated  as a decentralised, continual, and context-aware learning paradigm that departs from federated, distributed, and classic decentralised optimisation by treating nodes as persistent learners rather than interchangeable contributors to a shared objective. A high-level mathematical abstraction is introduced to separate local adaptation, opportunistic knowledge exchange, and context-conditioned integration, while remaining agnostic to specific model families or exchange mechanisms. Building on this abstraction, the paper examines system-level implications for communication, hardware heterogeneity, interoperability, trust, and governance, and discusses how resource sharing and interaction among nodes can give rise to continual intelligence across edge ecosystems.

\section{Background}
\label{sec:related_work}
\paragraph{\textbf{Pervasive and Context-Adaptive Edge Intelligence}}
Pervasive computing articulated an early vision of computation embedded seamlessly into everyday
environments, adapting to users and context rather than demanding explicit interaction
\cite{weiser1991computer}. Subsequent work demonstrated how multi-modal sensing and on-device
intelligence support context-aware adaptation across urban spaces, smart environments, and personal
devices, under constraints of energy, privacy, and intermittent connectivity
\cite{Parallelkanjo25,10.1145/3631618}.

Edge AI extends this vision by relocating inference and, increasingly, learning to the periphery of
the network. Hardware-efficient model architectures such as MobileBERT, combined with pruning,
quantisation, and distillation, enable transformer-grade models to operate on memory- and
energy-constrained devices \cite{sun2020mobilebert,molchanov2020pruning}. Complementary work on Tiny
Federated Learning shows that collaborative optimisation can be realised even on
microcontroller-class platforms through careful co-design of models and protocols
\cite{he2022tinyfl}. Mittal frames these developments as part of a broader transition from
cloud-centric analytics to increasingly autonomous edge computation.

Domain-specific deployments, including livestock behaviour recognition and environmental
monitoring, demonstrate that multi-modal learning pipelines can execute at the edge while meeting
strict latency and power budgets \cite{zhang2025multicore}. While this body of work establishes the
technical substrate for Node Learning, it typically retains either central orchestration or
fixed, application-specific coordination structures.

\paragraph{\textbf{Architectural Differences in Federated, Distributed, Collaborative, and Continuum Learning}}
Learning architectures for edge AI differ primarily in how coordination, objectives, and node roles
are structured. Federated Learning marked an early departure from centralised training by keeping
data local while enforcing a shared global objective through a coordinating server
\cite{kairouz2021federated}. Although effective in privacy-sensitive settings, this central point
of coordination constrains scalability and robustness under non-IID data, uneven participation,
and straggler effects \cite{baccour2022pervasive}.

Distributed learning follows a similar optimisation logic but assumes tighter coordination and
relatively stable infrastructure, prioritising efficiency over autonomy and treating nodes as
interchangeable workers contributing to a shared task. Removing the coordinating server leads to
decentralised learning, where aggregation is replaced by peer-to-peer interaction. Gossip and
neighbourhood-based protocols allow updates to propagate without global synchronisation, improving
resilience under dynamic connectivity \cite{xu2024aiot}. Wireless bilayer gossip further shows that
learning can persist under severe resource constraints in TinyML systems
\cite{bao2025bilayergossip}. However, despite architectural decentralisation, many such approaches
still assume a single shared optimisation objective and homogeneous node roles, effectively
replicating distributed optimisation without a server \cite{zangana2024gossip}.

Collaborative learning emerges when this assumption is relaxed. Nodes are no longer equivalent
contributors to a single model, but autonomous entities whose sensing, viewpoints, or capabilities
complement one another. Cooperation becomes selective and context-driven, and inference itself may
be collective. In edge perception, collaborative object detection demonstrates that exchanging
intermediate features or semantic hypotheses across neighbouring devices improves accuracy
\cite{richards2025edge}. Sparse cooperative perception further shows that selective communication
policies preserve these gains while respecting bandwidth and energy constraints
\cite{yuan2025sparsealign}. Here, accuracy improvements arise from functional complementarity and
information diversity rather than workload distribution or parameter averaging.

Interaction with cloud or fog layers remains possible but selective. Higher layers may support
escalation, synchronisation, or model distillation, such as refreshing frozen components or
sharing abstract knowledge across sites. Learning and decision-making remain anchored at the node,
while the continuum provides support when local capability or context is insufficient. Partial
offloading, progressive inference, and hierarchical aggregation balance task utility against
resource cost. This view aligns with work framing edge intelligence as the ``last mile'' of AI,
where cloud-scale models are adapted to operate under constrained and dynamic conditions
\cite{zhou2019edgeintelligence,baccour2022pervasive}. In domains such as smart manufacturing and
industrial sensing, edge devices exhibit persistent context, stable power supply, and task
specialisation, allowing them to function as long-lived loci of cognition rather than transient
participants.

\section{The Rise of Decentralised AI Toward Edge Intelligence}

\paragraph{\textbf{Towards Decentralised Intelligence}}
The emergence of decentralised intelligence reflects a broader shift in how computation, learning,
and decision-making are organised. Edge AI increasingly moves beyond relocating inference from the
cloud to enabling devices to interpret, learn, and act autonomously within the environments where
data are generated \cite{mittal2025evolution}. This shift has been enabled by advances in both model
design and embedded systems.

On-device learning has become feasible through compact, hardware-efficient architectures such as
MobileBERT \cite{sun2020mobilebert}, alongside pruning and quantisation techniques that reduce model
size and energy cost while preserving accuracy \cite{molchanov2020pruning}. Work on Tiny Federated
Learning further demonstrates that collaborative optimisation can operate within tight memory and
power budgets, even on microcontroller-class devices \cite{he2022tinyfl}. In parallel, advances in
embedded hardware—multi-core microcontrollers, low-power DSPs, and lightweight NPUs—support
concurrent sensing, inference, and adaptation under strict energy constraints
\cite{baccour2022pervasive}. Platforms combining MCUs with small TPUs have demonstrated multi-modal
TinyML pipelines in sub-watt regimes, including livestock behaviour monitoring
\cite{zhang2025multicore}.

Connectivity advances reinforce this decentralisation trend. Low-power wireless protocols,
including BLE~5.x, IEEE~802.15.4, and long-range technologies such as LoRaWAN, enable persistent
communication at minimal energy cost \cite{augustin2016lora,centenaro2016longrange}. LoRa-based
systems support kilometre-scale transmission with current consumption in the tens of milliamps,
making peer-level coordination feasible in wide-area deployments \cite{mroue2022lorawan}. Recent
microcontroller families integrate radios and AI acceleration on a single chip, forming unified
compute--communication substrates well suited to decentralised learning
\cite{stm32wl2024}.

\begin{figure}
    \centering
    \includegraphics[width=1\linewidth, height=0.3\linewidth ]{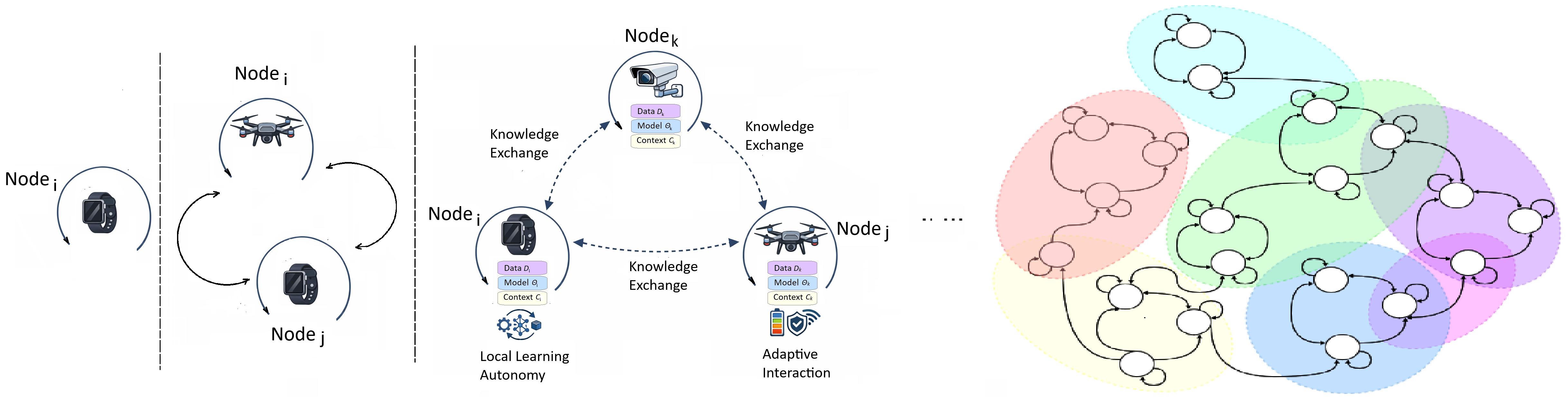}
    \caption{Node Learning from individual Edge AI nodes to opportunistic peer-to-peer collaboration and large-scale adaptive learning structures. Each node maintains local data, model state, and context, and exchanges learned knowledge with peers when beneficial. Overlapping collaboration regions enable knowledge diffusion and collective intelligence without central coordination or global aggregation.}
    \label{fig:NodeLearnigarchi}
\end{figure}

Early learning architectures such as Federated Learning represented an initial step by keeping
data local while coordinating optimisation centrally \cite{kairouz2021federated}. However, reliance
on a central aggregator introduces structural bottlenecks and limits robustness under non-IID data,
device heterogeneity, and dynamic participation \cite{baccour2022pervasive}. Increasingly, intelligence is distributed across a fabric of autonomous entities with persistent context and local learning capability, while higher layers are used selectively for synchronisation, escalation, or knowledge transfer \cite{zhou2019edgeintelligence,liu2025edgecloud}.
\paragraph{\textbf{Definition of Node Learning:}}
\textbf{Node Learning} is a decentralised and continual learning paradigm in which intelligence is
anchored at the level of individual nodes and scales through opportunistic interaction. Each node
operates as an autonomous learner, adapts a local learnable state from its own observations, and
engages with peers at varying scales—from isolated operation, to pairwise exchange, to large and
dynamic populations—without assuming persistent coordination or global synchronisation (see figure ~\ref{fig:NodeLearnigarchi}).
\textbf{Single-node learning.}
Consider first an isolated node $i$. The node maintains a learnable state $\theta_i \in \Theta_i$
and adapts it using local data and context:
\begin{equation}
\min_{\theta_i} \; F_i(\theta_i)
= \mathbb{E}_{(x,y)\sim D_i}\!\left[\ell(\theta_i; x, y, c_i)\right],
\end{equation}
where $D_i$ denotes non-IID local data and $c_i$ captures contextual factors such as energy,
connectivity, sensing modality, and mobility. Local adaptation proceeds via an abstract update rule
\begin{equation}
\theta_i^{t+1} = \mathcal{U}_i\!\left(\theta_i^{t}, D_i, c_i\right),
\end{equation}
without reliance on external communication.

\textbf{Pairwise and small-group interaction.}
When one or more peers become available, the node may engage in selective collaboration. For a
time-varying neighbour set $\mathcal{N}_i(t)$ of cardinality $|\mathcal{N}_i(t)| \geq 1$, knowledge
exchange is expressed through a context-aware merge operator
\begin{equation}
\theta_i^{t+1} \leftarrow
\mathcal{M}_i\!\left(
\theta_i^{t+1},
\{\theta_j^{t}\}_{j\in\mathcal{N}_i(t)},
c_i, c_j
\right),
\end{equation}
where interaction may involve a single peer, a small cluster, or a transient group. The operator
$\mathcal{M}_i(\cdot)$ abstracts the form of exchange and may correspond to feature sharing, partial
model updates, distillation, or other transfer mechanisms.

\textbf{Population-level interaction.}
At scale, Node Learning extends naturally to large and evolving populations, potentially comprising
millions of nodes. There might not be a global objective explicitly optimised. In general global behaviour emerges
from repeated local adaptation and opportunistic interaction. 

Node Learning also allows nodes to pursue \textbf{distinct objectives} while still benefiting from interaction, nodes may share transferable components of their learnable state.
While task-specific parameters remain private and locally optimised. For example, a roadside camera trained for traffic flow estimation may share motion embeddings with a mobile device trained for pedestrian tracking. Although the objectives differ, the exchanged representations encode shared spatio-temporal structure that improves both tasks.
Learning gains arise from exploiting common structure in the
environment rather than from objective alignment, allowing heterogeneous nodes to collaborate effectively while preserving autonomy.

\textbf{It's More Than Distributed AI}:
When beneficial, nodes may distribute computation, approximating a shared objective
$\min_{\theta}\sum_i F_i(\theta)$, or collaborate through peer consensus
$\theta_i \approx \theta_j$. Node Learning subsumes both by making distribution and
collaboration optional, realised through context-driven interactions
$\theta_i^{t+1} \leftarrow \mathcal{M}_i(\cdot)$ rather than enforced by aggregation.
\paragraph{\textbf{Heterogeneous and Homogeneous Nodes}}

Within Node Learning, both heterogeneous and homogeneous regimes are supported, encompassing nodes
that differ in memory capacity, processing power, energy availability, sensing modality, and
communication capability. Such networks include wearables, drones, environmental sensors, and
industrial robots operating within shared physical spaces. Heterogeneity introduces functional
diversity: for example, a ground robot may optimise spatial mapping, an aerial drone may focus on
visual coverage, and a stationary sensor may capture microclimatic signals. Each node $i$ therefore
optimises a task- and context-specific objective
\begin{equation}
\min_{\theta_i} F_i(\theta_i \mid c_i),
\end{equation}
where the loss $F_i$ and context $c_i$ vary across nodes. Collectively, the network forms a
composite representation of the environment through complementary learning processes. Hossain
\emph{et al.} describe this behaviour as \emph{functional complementarity}, in which nodes with
distinct roles contribute to a broader inference process without uniform participation
\cite{hossain2023collaborative}.

Heterogeneous networks improve adaptability and robustness but complicate coordination. Nodes must
adjust learning rates, exchanged representations, and collaboration frequency according to their
capabilities. This behaviour can be captured through context-weighted interaction,
\begin{equation}
\theta_i^{t+1} \leftarrow
\mathcal{M}_i\!\left(
\theta_i^{t+1},
\{\phi_j^{t}\}_{j\in\mathcal{N}_i(t)},
c_i,c_j
\right),
\end{equation}
where $\phi_j$ denotes transferable knowledge and higher-capacity nodes may temporarily assume
coordination roles. Such roles are opportunistic rather than structural. Lower-capacity nodes may
offload computation or rely on distilled knowledge from nearby peers while retaining autonomy.

\begin{figure}
    \includegraphics[width=1\linewidth]{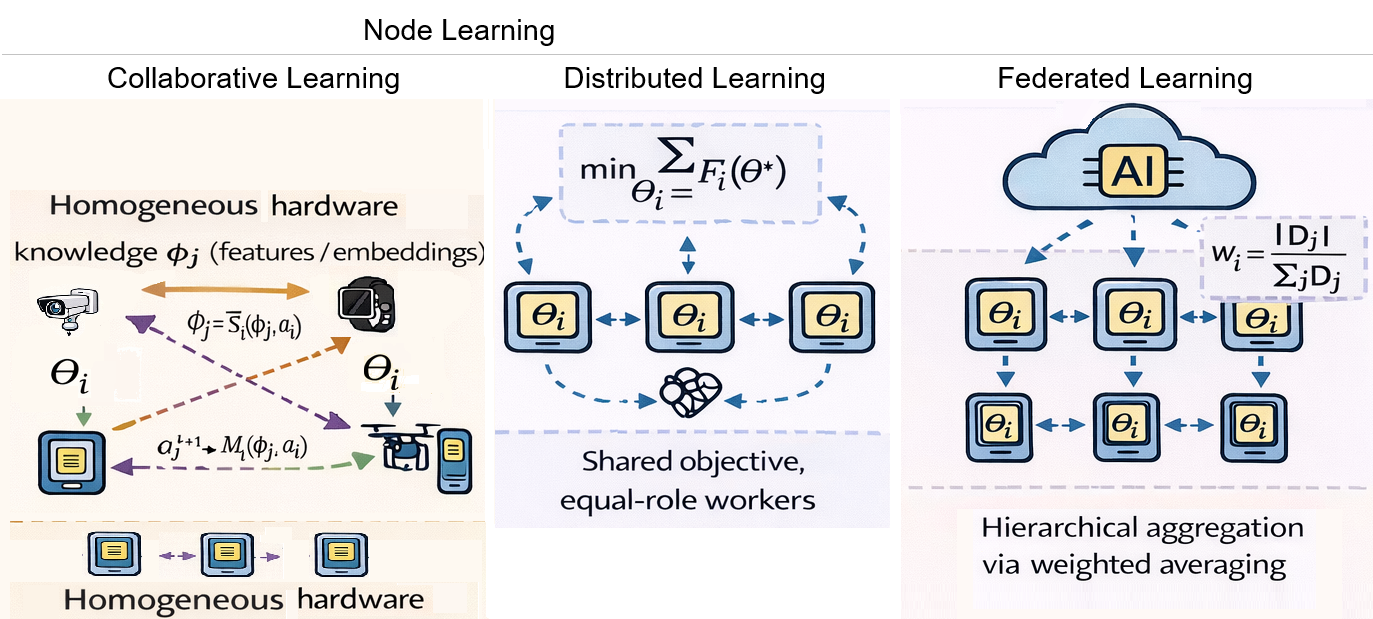}
    \caption {Conceptual comparison of collaborative, distributed, and federated learning. Collaborative learning shows context-driven peer exchange of learned representations without shared objectives or averaging; distributed learning optimises a common objective with equal participation; federated learning uses server-mediated weighted aggregation. Interaction patterns are illustrative and subject to design and application choices.}
    \label{fig:collabdistributefed}
\end{figure}

In contrast, homogeneous networks comprise nodes with identical specifications and learning
objectives, such that $F_i \equiv F$ and $\Theta_i \equiv \Theta$ for all $i$. Coordination is
simplified and often relies on deterministic consensus mechanisms,
\begin{equation}
\theta^{t+1} = \sum_{i=1}^{N} w_i \theta_i^{t+1},
\end{equation}
including weighted averaging or majority voting. Such systems are effective for tasks such as
distributed fault detection using arrays of identical sensors, where redundancy improves
reliability \cite{zangana2024gossip}. However, homogeneity limits adaptability in multi-modal or
dynamic environments, as identical perspectives cannot capture diverse contextual cues. See figure ~\ref{fig:collabdistributefed}.

\subsection{\textbf{Data Dynamics and Contextual Adaptation}}
Classical machine learning commonly assumes that data samples are independent and identically
distributed (IID). For a dataset $\{(x_i,y_i)\}_{i=1}^{N}$, this implies
\begin{equation}
p(x_1,y_1) = p(x_2,y_2) = \cdots = p(x_N,y_N),
\end{equation}
with $(x_i,y_i)$ independent of $(x_j,y_j)$ for $i \neq j$. This assumption underpins convergence
guarantees in centralised training, where data share uniform statistical properties.

In decentralised and edge intelligence settings, this assumption rarely holds. As defined earlier,
each node optimises a local objective $F_i(\theta)$ induced by its data distribution and context.
In practice, these distributions are both non-IID across nodes and non-stationary over time. Each
node observes data shaped by its environment, sensing modality, user behaviour, and operational
conditions, leading to time-varying sampling processes
\begin{equation}
(x,y) \sim p_i(x,y \mid c_i(t)),
\end{equation}
where $c_i(t)$ captures contextual factors such as location, energy state, and environmental
conditions.
Such statistical heterogeneity causes local optimisation trajectories to diverge, degrading the
effectiveness of naïve aggregation schemes and static consensus. In Node Learning, this diversity is not treated as a deviation from an idealised setting, but as a defining characteristic. Nodes perceive different subsets of the world, operate at different sampling rates, and experience distribution shift driven by factors such as weather, illumination, mobility, or crowd dynamics.

Nodes adjust their learning dynamics based on contextual relevance and resource availability,
\begin{equation}
\theta_i^{t+1} = \theta_i^{t} - \eta\,\omega_i(c_i(t))\,\nabla F_i(\theta_i^{t}),
\end{equation}
where $\omega_i(c_i(t))$ modulates the influence of new observations. This allows nodes to adapt rapidly to local events while limiting destabilising drift. Context-aware mechanisms, such as salience-weighted updates, demonstrate how local relevance can guide adaptation without sacrificing stability \cite{parisi2019continual}. In practice, this enables personalised learning—such as wearable devices adapting health thresholds—while maintaining latent alignment through selective peer interaction. 
This allows Node Learning to represent:

\vspace{-0.7\baselineskip}
\begin{itemize}
  \item Temporal evolution and scene change;
  \item Sensor and modality variation;
  \item Event-based triggers for learning updates.
\end{itemize}
\vspace{-0.7\baselineskip}

\subsection{Continuum Computing across Edges}

In Node Learning, networked edges are treated as fluid layers between which computation and learning migrate dynamically, introducing   two key properties—\emph{plasticity} and \emph{elasticity}—that allow learning
to persist across hierarchical boundaries \cite{liu2025edgecloud}. Plasticity captures the ability of models to adapt over time, while elasticity enables computation and communication to scale vertically and laterally across the continuum.

Formally, let $\Theta_{\text{edge}}$, $\Theta_{\text{fog}}$, and $\Theta_{\text{cloud}}$ denote
learnable states at different layers of the continuum. Cross-layer adaptation in Node Learning is
captured by
\begin{equation}
\Theta_{\text{edge}}^{t+1} \leftarrow 
\phi\!\left(
\Theta_{\text{edge}}^{t},
\Theta_{\text{fog}}^{t},
\Theta_{\text{cloud}}^{t},
c_i(t)
\right),
\end{equation}
where $c_i(t)$ encodes contextual factors such as bandwidth, relevance, latency, and energy.

At the node level, learning evolves through the interplay of elasticity and plasticity. Plasticity
is expressed as local adaptation driven by non-IID observations,
\begin{equation}
\theta_i^{t+1} = \theta_i^{t} + \Delta_i(t),
\end{equation}
where $\Delta_i(t)$ reflects task- and context-dependent updates. Elasticity governs the evolution
of the collaboration neighbourhood $\mathcal{N}_i(t)$ and the associated interaction weights
$w_{ij}(t)$, determining when and how nodes engage with peers.

\subsection{Collaborative Accuracy through Shared Resources}

Node Learning allows a population of constrained devices to act collectively as a system with
greater effective capability than any individual node. Rather than scaling a single model, nodes
share resources opportunistically, enabling distributed memory, computation, and perception. For
example, in a social gathering, multiple wearables—each limited to a few megabytes of memory—can
collaborate to support tasks that exceed individual capacity. If node $i$ maintains a compact
local representation of size $m_i \approx 2$\,MB, the effective accessible capacity becomes
\begin{equation}
M_{\mathrm{eff}} \approx \sum_{j \in \mathcal{N}_i} m_j,
\end{equation}
which can reach tens of megabytes through decentralised sharing of memory and compute, without
central aggregation.

Accuracy improvements arise from the structure of collaboration rather than raw scale. Nodes with
different sensing modalities contribute complementary evidence, while exchanged representations
or intermediate features act as implicit regularisers under non-IID data. Observations of rare or
extreme events by individual nodes can be propagated selectively, improving population-level
robustness and reducing overfitting to local conditions.

Resource sharing in Node Learning spans multiple dimensions and adapts dynamically to context, as
summarised in Table~\ref{tab:resource_sharing}. Compute sharing enables partial offloading or
distributed inference when local load or thermal limits are reached. Memory sharing allows nodes
to externalise embeddings, replay buffers, or heavy model components onto more capable peers.
Sensing resources are combined through cross-modal validation and calibration, improving inference
under occlusion or noise. Communication resources are exploited through opportunistic relays that
bridge heterogeneous links, while energy-aware role rotation balances participation and prolongs
system lifetime.

Together, these mechanisms allow Node Learning systems to improve accuracy and capability not by
centralising models or data, but by composing distributed resources into a flexible, context-aware
collective.

\begin{table}[t]
\centering
\caption{Resource sharing mechanisms in Node Learning.}
\label{tab:resource_sharing}
\renewcommand{\arraystretch}{1.15}
\begin{tabular}{@{}p{2.5cm}p{4cm}p{4.5cm}@{}}
\toprule
\textbf{Resource} & \textbf{Mechanism} & \textbf{Effect} \\ \midrule
Compute & Distributed inference, partial offload & Reduces local overload and latency \\
Memory & Shared embeddings, adapters & Increases effective model capacity \\
Sensing & Cross-modal fusion and calibration & Improves inference accuracy \\
Communication & Opportunistic relays & Extends reach, reduces energy cost \\
Energy & Role rotation, duty-cycle adaptation & Prolongs system lifetime \\ 
\bottomrule
\end{tabular}
\end{table}

\subsection{Wireless Exchange, Clustering, and Opportunistic Interaction}
\paragraph{\textbf{Wireless Exchange of Learned State}}

In dynamic edge environments such as vehicular systems, mobile sensing platforms, or drone swarms,
wireless connectivity is inherently intermittent, asymmetric, and spatially correlated. Nodes may
encounter one another briefly and unpredictably, with highly variable link quality and contact
duration. In such settings, decentralised learning removes the coordinating entity altogether.
Gossip-based learning and decentralised stochastic gradient descent (D-SGD) have shown that direct
peer-to-peer exchange can achieve competitive convergence while improving robustness to node
dropout, delayed updates, and link failures. Learning is reframed as a diffusion process, where
updates propagate locally and asynchronously until approximate alignment emerges, without global
synchronisation or fixed communication rounds.

\begin{figure}
    \centering
    \includegraphics[width=1\linewidth]{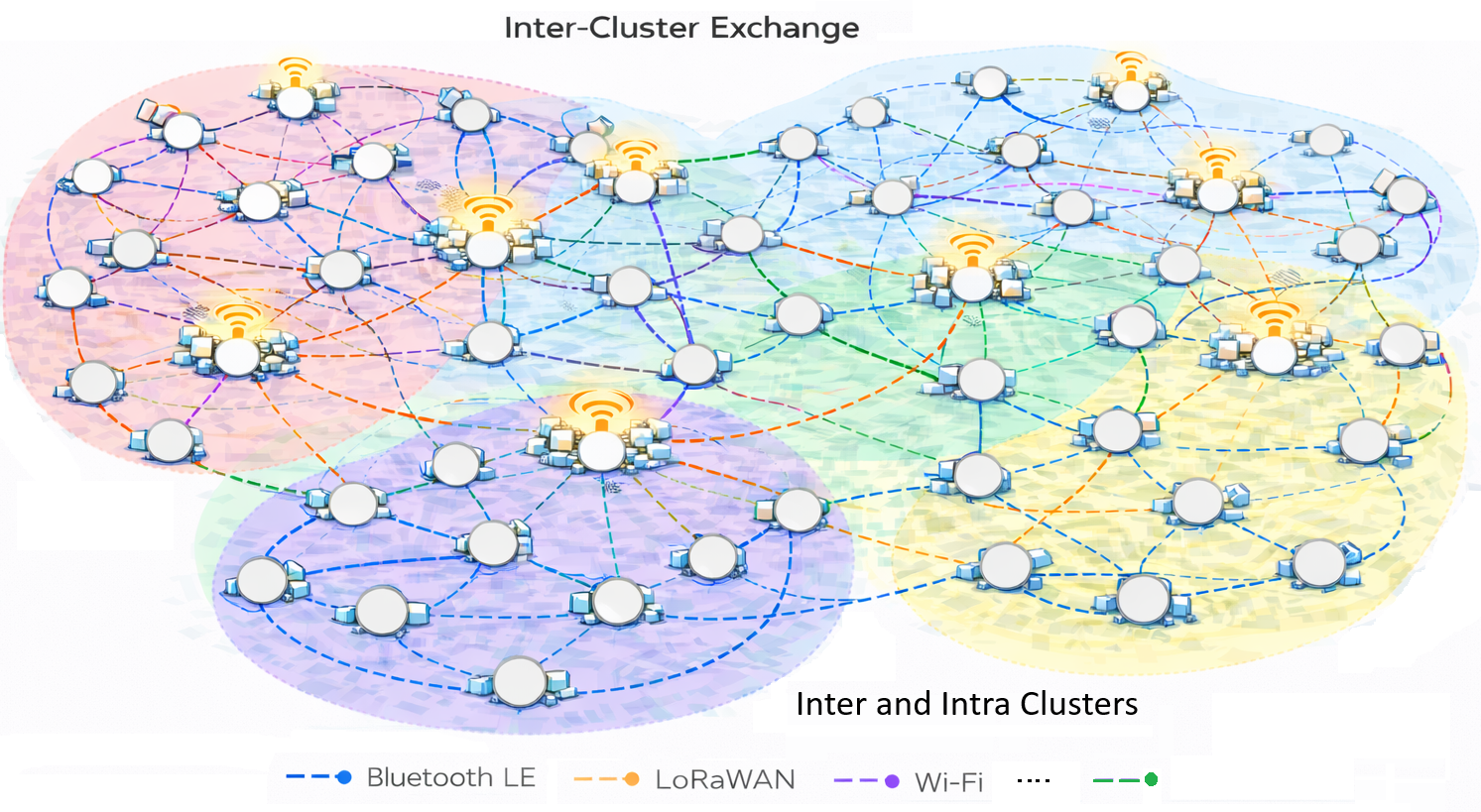}
    \caption{Wireless exchange of learned state under opportunistic clustering.}
    \label{fig:wirlessclusters}
\end{figure}

Node Learning extends decentralised communication by explicitly supporting \emph{opportunistic
collaboration}. Wireless encounters provide the mechanism for interaction, clustering defines an interaction strategy, which modulates the effect of exchanged knowledge on local learning.
Nodes form temporary coalitions based on proximity, channel quality, energy state,
trust, or task relevance, and dissolve them as conditions change. These coalitions are ephemeral
rather than statically clustered, and interaction is triggered by opportunity rather than schedule.
Communication is therefore sparse, asymmetric, and context-driven. While conceptually related to
Delay-Tolerant Federated Learning, Node Learning removes residual hierarchical assumptions: there is
no designated aggregator, no obligation to participate in rounds, and no requirement for eventual
global model synchronisation. Nodes exchange only locally relevant knowledge at the time of contact,
using lightweight wireless technologies such as Bluetooth Low Energy, LoRaWAN, Wi-Fi Direct, or ad
hoc vehicular links (see Figure~\ref{fig:wirlessclusters}).

Clustering can nevertheless serve as a lightweight structuring mechanism in large-scale
deployments, without imposing global coordination. Nodes may form time-varying groups
$C_k \subset \{1,\ldots,N\}$ based on proximity, context, or task similarity. Within a cluster,
interaction follows the same merge operator defined earlier, restricted to peers in $C_k$. Exchange
across clusters is less frequent and relies on compressed or distilled summaries that capture
cluster-level knowledge,
\begin{equation}
\theta_i^{t+1} \leftarrow
\mathcal{M}_i\!\left(
\theta_i^{t},
\{\phi_\ell^{t}\}_{\ell \in C_k,\, \ell \neq i}
\right).
\end{equation}

In such configurations, roles remain opportunistic rather than structural:
(i) higher-capability nodes may temporarily act as local coordinators;
(ii) inter-cluster exchange prioritises compressed or low-rank representations;
(iii) reputation- or confidence-weighted policies bias transmission toward reliable contributors.
Crucially, these mechanisms structure interaction without reintroducing persistent aggregation or
central control.

\subsection{Hardware and Evaluation in Node Learning}

\paragraph{\textbf{Hardware Constraints and Opportunities}}

Node Learning places stringent demands on hardware by pushing continual learning directly onto
devices operating under tight power, memory, and latency constraints. Unlike conventional Edge AI,
which often relies on inference-only execution, Node Learning requires persistent adaptation on
ultra-low-power platforms, including ARM Cortex-M4/M7 microcontrollers and emerging RISC-V--based
NPUs. In many deployments, learning must proceed autonomously within milliwatt-scale power
envelopes and kilobyte-to-megabyte memory budgets.

Recent hardware developments increasingly support this shift. TinyML platforms integrate DSP units
that enable efficient feature extraction and incremental updates under strict energy constraints.
Neuromorphic processors support event-driven computation aligned with sparse, asynchronous sensor
streams, making them well suited to continual adaptation without dense clocked execution. RISC-V
NPUs expose programmable data paths and lightweight accelerators that support partial retraining,
adapter updates, and decentralised inference pipelines. Together, these architectures extend
learning capability beyond edge servers to deeply embedded devices.

Crucially, Node Learning does not assume uniform hardware capability. Instead, it exploits
heterogeneity: lightweight nodes prioritise local adaptation and compact representations, while
more capable devices temporarily amplify collective capability through distillation, caching, or
selective coordination—without becoming structural points of control.
\begin{table}[t]
\centering
\caption{Analytical reflection on hardware platforms for Node Learning.}
\label{tab:hardware_reflection}
\renewcommand{\arraystretch}{1.15}
\begin{tabular}{@{}p{2.8cm}p{3.2cm}p{4.5cm}@{}}
\toprule
\textbf{Hardware Class} & \textbf{Key Capability} & \textbf{Implication for Node Learning} \\ \midrule
MCUs (TinyML) & Ultra-low power, tight memory &
Enable continual local adaptation and feature-level exchange; require aggressive pruning,
quantisation, and sparse updates. \\
Edge NPUs (RISC-V, TPU-lite) & Moderate compute, programmable pipelines &
Support partial retraining, adapter updates, and distributed inference without cloud offload. \\
Neuromorphic chips & Event-driven, sparse computation &
Well suited for asynchronous learning and intermittent sensing under extreme energy constraints. \\
Edge servers / Cloudlets & High compute, stable power &
Act as transient capability amplifiers for distillation or caching, without becoming structural
coordinators. \\
Cloud & Large-scale optimisation &
Used selectively for pretraining or periodic alignment, not for routine orchestration. \\
\bottomrule
\end{tabular}
\end{table}

\begin{table}[t]
\centering
\caption{Testbeds and benchmarking dimensions for Node Learning evaluation.}
\label{tab:testbeds_benchmarks}
\renewcommand{\arraystretch}{1.15}
\begin{tabular}{@{}p{3cm}p{3.5cm}p{4cm}@{}}
\toprule
\textbf{Testbed Type} & \textbf{Typical Setup} & \textbf{Benchmark Focus} \\ \midrule
Homogeneous clusters & Identical SBCs or MCUs (10--50 nodes) &
Algorithmic stability, baseline energy, and accuracy behaviour. \\
Heterogeneous deployments & Mix of MCUs, NPUs, edge servers &
Adaptability to resource imbalance, role switching, collaboration efficiency. \\
Wireless mesh networks & Wi-Fi, BLE, LoRa connectivity &
Robustness under latency, packet loss, and intermittent links. \\
Simulation frameworks & EdgeSimPy, FederatedScope &
Scalability, mobility, and large-population dynamics. \\
Real-world pilots & Wearables, drones, urban sensors &
Context adaptation, resilience, and long-term autonomy. \\
\bottomrule
\end{tabular}
\end{table}

\paragraph{\textbf{Evaluation, Benchmarking, and Experimental Setups}}

Evaluating Node Learning requires metrics that go beyond conventional Edge AI measures such as
accuracy and latency. Key performance indicators include \emph{energy per iteration} (EPI),
capturing the cost of continual adaptation; \emph{collaboration efficiency} (CE), defined as
accuracy improvement per byte exchanged; \emph{adaptation latency} (AL), measuring responsiveness
to context or distribution shift; and \emph{resilience ratio} (RR), quantifying performance
degradation under node dropout or communication loss.

While simulation environments such as EdgeSimPy, FederatedScope, and SimFlow remain valuable for
scalability analysis, Node Learning evaluation increasingly requires empirical validation. Typical
experimental setups combine controlled and heterogeneous deployments. Homogeneous clusters are used
to isolate algorithmic behaviour, while heterogeneous deployments mix microcontrollers, NPUs, and
edge servers to assess adaptability under resource imbalance and dynamic role allocation.
Cross-domain datasets spanning audio, vision, and motion are commonly employed to evaluate
multi-modal learning under non-IID conditions.

A representative configuration involves 10--100 nodes connected via Wi-Fi mesh, BLE, or low-power
long-range links such as LoRa. Nodes perform continual on-device learning and engage in
opportunistic peer interaction without synchronous aggregation. Such setups allow direct
measurement of energy–accuracy trade-offs, communication efficiency, and robustness under realistic
wireless conditions.
A minimal evaluation of Node Learning therefore needs to include non-IID and evolving data, intermittent or asymmetric connectivity, and the absence of a persistent coordinating entity; evaluations that violate these assumptions do not exercise the defining properties of the paradigm.

\section{Systemic Considerations}

The trajectory of AI is increasingly shaped by system-level coherence rather than isolated technical
advances. Hardware capability, learning dynamics, communication, and governance now interact tightly,
particularly outside the data centre. Node Learning frames intelligence as an emergent property of
interacting entities, anchored at the edge and shaped by context, rather than delegated to central
infrastructure. This reflects a shift from connectivity-driven systems to cognition-driven networks,
in which devices learn, adapt, and contribute locally.

Earlier paradigms—including centralised cloud computing, data-centre AI, and Federated Learning—
connected devices to remote or hierarchical intelligence. Node Learning instead treats devices as
persistent sources of intelligence. Knowledge flows opportunistically and contextually, replacing
linear data pipelines with adaptive peer interaction. As noted by Yu \emph{et al.}, decentralised
collaborative learning reshapes communication into dynamic trust-weighted networks, where alignment
emerges through interaction rather than orchestration \cite{yu2023trustworthy}.

\paragraph{\textbf{Sustainability and Cost Reduction}}

Decentralised intelligence offers concrete sustainability benefits by reducing dependence on
continuous cloud interaction. Local learning and selective collaboration lower communication
overhead, reduce backhaul traffic, and decrease the energy footprint associated with large-scale data
centres. Empirical studies show that adaptive local updates under non-IID conditions can deliver
significant energy savings while maintaining, or improving, task performance.

By contrast, centralised data centres incur substantial capital and operational costs due to cooling,
networking, redundancy, and overprovisioned compute. Node Learning reduces repeated data transmission,
limits redundant computation, and mitigates peak demand on shared infrastructure. Decentralised
resource pooling allows populations of constrained devices to collectively approach, or exceed, the
capability of a single central model, while avoiding the financial, latency, and environmental costs
of cloud-centric orchestration.

\paragraph{\textbf{Ethics, Accountability, Trust, and Misuse}}

As autonomy increases and learning persists in open environments, responsibility in Node Learning
becomes inherently distributed. Accountability can no longer be enforced through a single control
point, yet existing ethical and governance mechanisms do not translate cleanly to this setting.
Node Learning systems often operate on severely constrained hardware, with limited memory,
computation, and temporal context, making even lightweight responsible-AI or cryptographic mechanisms
difficult to sustain.

Ethical operation therefore relies on constraints rather than guarantees. Data minimalism, limited
state persistence, and partial decision traceability guide system design under resource limits.
Lightweight explainability tools, such as TinyML variants of SHAP or Micro-LIME, offer bounded local
transparency, but their applicability remains situational.

Trust in Node Learning cannot depend on heavy infrastructure, long histories, or global reputation.
Instead, it is embedded into learning itself as a short-horizon, adaptive signal informed by recent
reliability, contextual relevance, and observed utility. Nodes decide when and with whom to
collaborate based on current evidence. In a smart-city setting, for example, a roadside sensor may temporarily prioritise updates from nearby cameras during peak traffic while discounting stale or
low-confidence contributors.
Decentralised trust graphs, in which edge weights evolve
with collaborative performance, enable trust-aware routing of learned state and improve robustness
under unreliable participation \cite{shao2023trustednetwork}. Reputation-aware learning stabilises collaboration without reintroducing central control.

A key risk is malicious misuse or unintended agentic behaviour. Autonomous nodes that adapt
continuously and exchange updates may drift toward undesirable behaviour under partial observability
and short memory. While deliberate misuse is possible, unintentional misalignment is often more
concerning: locally rational updates that collectively reinforce bias, amplify spurious
correlations, or degrade system behaviour. In such settings, frequent adaptation can become a
liability rather than an asset.

Existing work on trusted consensus, decentralised trust graphs, and incentive mechanisms provides
useful foundations \cite{wang2021trusted,shao2023trustednetwork}, but these approaches were not
designed for highly constrained, agent-like learners with ephemeral state. Node Learning therefore
requires trust-by-design rather than trust-by-retrofit, with governance focused on bounding harm
rather than enforcing perfect compliance.

\paragraph{\textbf{Security}}

Peer-to-peer learning introduces new attack surfaces, including model poisoning, inference leakage
through shared representations, and trust degradation in open networks. Techniques such as
homomorphic encryption, secure aggregation, anomaly detection, and local differential privacy can
reduce risk, but their overhead remains prohibitive for many edge devices. Security mechanisms must
therefore be selective and context-dependent rather than uniformly enforced.

Decentralisation also reframes ownership and decision authority. Each node retains control over its
local model state, while collective behaviour emerges from repeated local interaction rather than
explicit aggregation. This layered structure—local autonomy, peer alignment, and population-level
emergence—avoids single points of failure while preserving accountability through local traceability
\cite{xu2024aiot}.

\paragraph{\textbf{Interoperability}}

Heterogeneity in hardware, operating systems, communication protocols, and model architectures
presents a parallel challenge. Sustainable Node Learning requires abstraction layers that decouple
learning logic from platform-specific constraints. Practical enablers include common APIs for
learned-state exchange, cross-architecture compilation frameworks such as TVM and ONNX Runtime
Mobile, and translation layers that map representations across model families.

Beyond technical compatibility, interoperability intersects with regulation and governance.
Standards bodies including IEEE, ETSI, and ISO are exploring decentralised AI standards, trustworthy
Edge AI frameworks, and data locality compliance mechanisms.

\section{Opportunities and Future Directions}

Advances in TinyML, model compression, in-memory processing, neuromorphic computing, and programmable
NPUs continue to expand the feasibility of continual on-device learning. Algorithmic integration
with graph learning, swarm intelligence, and representation learning supports scalable decentralised
adaptation, where knowledge propagates through interaction rather than coordination. In parallel,
standardisation efforts are emerging around learned-state exchange, energy-aware scheduling, and
edge–cloud interoperability.

\paragraph{Node Language Models and Generative AI}

Recent progress in small and medium-scale language models extends Node Learning beyond perception
into local reasoning, planning, and semantic interpretation. Foundational models are typically
deployed in frozen or partially frozen form, with adaptation achieved through lightweight mechanisms
such as adapters, low-rank updates, or prompt conditioning. This avoids full retraining while
supporting continuous contextual specialisation on constrained devices.

Adaptation is decentralised. Nodes refine behaviour through prompt evolution, adapter tuning, or
task-specific heads, and exchange distilled knowledge—such as embeddings, prompts, or confidence
signals—rather than full model weights. Resource sharing enables generative functionality beyond
individual capacity, with compute-intensive operations or memory-heavy components distributed across
neighbourhoods. Over time, populations of specialised generative models emerge, sharing structure
without sacrificing autonomy.

\paragraph{Applications}

In agriculture, sensors, drones, and machinery collaborate to manage livestock
\cite{zhang2025multicore}, optimise irrigation, and detect crop stress. In urban and rural
infrastructure, decentralised learning supports traffic management \cite{fu2025trajaware}, energy
optimisation, and environmental monitoring under intermittent connectivity. In disaster response,
wearables, robots, and aerial platforms form ad hoc learning networks that share situational
awareness without relying on intact infrastructure.

Healthcare and wellbeing are equally important. Wearables and medical devices collaborate to learn
personalised health models, detect anomalies, and share high-level insights while preserving
privacy. Similar principles apply to security and defence, where resilient,
infrastructure-independent intelligence is essential under adversarial conditions.

\paragraph{\textbf{Taking Advantage of Novel Compute}}

Node Learning aligns naturally with emerging non-von Neumann computing paradigms. In-memory and
near-memory computing reduce data movement, enabling low-latency, energy-efficient adaptation.
Neuromorphic architectures support event-driven learning under sparse sensing. Analogue and
mixed-signal accelerators, memristive and phase-change devices, spintronics, and approximate
computing exploit physical dynamics to reduce power consumption, aligning with the tolerance of
learning algorithms to noise.

Open, programmable architectures particularly RISC-V NPUs enable custom learning primitives and
accelerate hardware algorithm co-design. Advances in sensing, including multimodal, nano-scale, and
bio-inspired sensors, expand the scope of local learning, while on-sensor processing compresses the
learning pipeline further. Progress in power systems, including energy harvesting and low-power
storage, supports long-lived autonomous operation. In parallel, learning-aware communication
protocols prioritise semantic updates over raw data, improving efficiency across heterogeneous
networks.
Collectively, these architectures reduce the marginal cost of adaptation—not just inference—making continual, decentralised learning feasible below the watt scale for the first time.
\section{Final Remarks}
The growing cost of centralised intelligence in data movement, energy consumption, latency, and infrastructure signals a shift in how learning systems are organised. Intelligence is increasingly distributed across devices that sense, adapt, and cooperate within their environments. Node Learning captures this shift by framing intelligence as a decentralised, cooperative process rooted in persistent local learning and opportunistic interaction.

Node Learning does not replace existing approaches but reframes them. Concepts from federated learning, distributed optimisation, collaborative perception, and edge intelligence appear as operational regimes within a broader decentralised landscape, distinguished by how coordination, objectives, and interaction are constrained. What distinguishes Node Learning is not a specific algorithm or protocol, but an architectural perspective in which learning persists and propagates without assuming stable infrastructure or central authority.

As edge ecosystems grow in scale and diversity, this perspective provides a way to reason about intelligence that is resilient, context-aware, and embedded in the physical world—an intelligence
that evolves through cooperation among many learning entities rather than execution at a single point.
The value of Node Learning lies in providing a unifying frame that makes previously fragmented design choices comparable within a single conceptual space.

\bibliographystyle{ACM-Reference-Format}
\bibliography{references}

\end{document}